%% file: main.tex
\documentclass{article}
\usepackage{iclr2027_conference,times}
\input{math_commands.tex}

\usepackage[hyperfootnotes=false,unicode,hidelinks]{hyperref}
\usepackage{url}
\usepackage{graphicx,booktabs,amssymb}

\iclrfinalcopy
\newcommand{\authmark}[1]{\textsuperscript{#1}}
\title{HoloNav-0: Specialist Foundation Models as Tools for Generalist Navigation}
\author{%
  \textbf{Pengna Li\authmark{1*\textdagger}, Kangyi Wu\authmark{1*},
  Yucheng Wang\authmark{2\textdaggerdbl}, Xiaolin Zhou\authmark{2}, Kailin Lyu\authmark{3}}\\
  \textbf{Xihan Ma\authmark{2}, Yance Fang\authmark{4}, Zhizhong Su\authmark{2},
  Nanning Zheng\authmark{1\textsection}}\\
  \normalfont\authmark{1}Institute of Artificial Intelligence and Robotics, Xi'an Jiaotong University
  \normalfont\authmark{2}Horizon Robotics\\
  \normalfont\authmark{3}Institute of Automation, Chinese Academy of Sciences
  \normalfont\authmark{4}Renmin University of China}
\hypersetup{
  pdftitle={HoloNav-0: Specialist Foundation Models as Tools for Generalist Navigation},
  pdfauthor={Pengna Li, Kangyi Wu, Yucheng Wang, Xiaolin Zhou, Kailin Lyu, Xihan Ma, Yance Fang, Zhizhong Su, Nanning Zheng},
  pdfsubject={Generalist embodied navigation with a shared policy and specialist tools}
}
\begin{document}
\maketitle
\lhead{Preprint}
\begingroup
\renewcommand{\thefootnote}{\fnsymbol{footnote}}
\footnotetext[1]{Pengna Li and Kangyi Wu contributed equally.}
\footnotetext[2]{Pengna Li interned at Horizon Robotics during this work. Email: \mbox{\href{mailto:sauerfisch@stu.xjtu.edu.cn}{\nolinkurl{sauerfisch@stu.xjtu.edu.cn}}}.}
\footnotetext[3]{Project leader: Yucheng Wang.}
\footnotetext[4]{Corresponding author: Nanning Zheng.}
\endgroup

\begin{abstract}
Generalist navigation requires diverse perceptual and reasoning capabilities
to support tasks such as instruction following and object search. Recent
generalist navigation models typically learn these capabilities within an
end-to-end policy, so adding or improving individual capabilities can require
further training of the navigation model. We argue that a navigation decision
model need not master every capability itself; it can instead use
independently pretrained specialist models as external tools to expand
navigation capabilities without retraining the decision model. We present HoloNav-0, a unified navigation architecture that combines generalist learning with modular extensibility. We train a single decision policy jointly for route following and object search and use specialist foundation models as external tools that can be replaced independently without retraining the policy. Perception tools recover missing depth
and reconstruct panoramic context, foresight tools generate future
observations conditioned on candidate moves, and verification tools ground
targets to validate the decision model's stopping proposals. This separation
offers two benefits. First, compatible tool backends can be replaced or
upgraded while keeping the decision model fixed, providing a path to
benefiting from advances in specialist models. Second, perception tools enable
one checkpoint to operate across panoramic and forward-facing RGB and
RGB-D configurations without sensor-specific fine-tuning.
HoloNav-0 achieves competitive performance in both vision-and-language
navigation and object-goal navigation with one shared decision model.
Specifically, it attains success rates of 71.6\% on R2R-CE val-unseen
and 56.7\% on HM3D-OVON val-unseen.
\end{abstract}

\section{Introduction}
\label{sec:intro}

\begin{figure}[t]
    \centering
    \includegraphics[width=\linewidth,pagebox=cropbox]
    {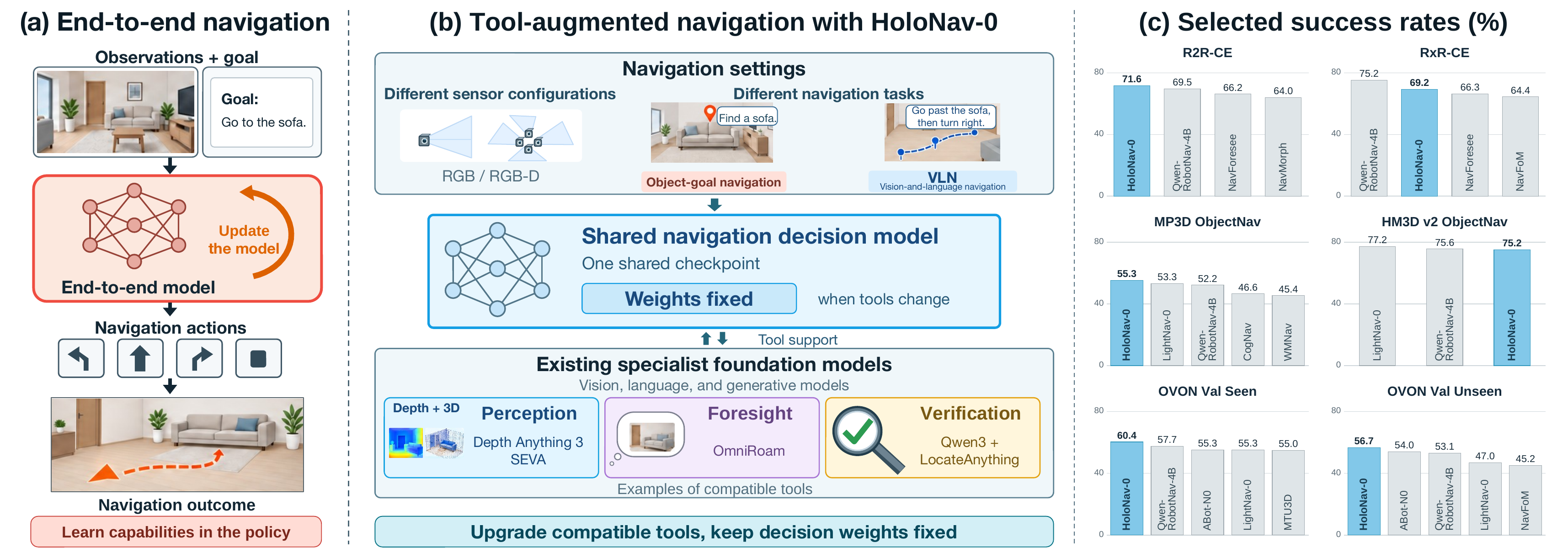}
    \caption{\textbf{Method and performance comparisons.}
    (a) End-to-end navigation.
    (b) HoloNav-0 combines a shared navigation decision model with external
    tools, allowing compatible tools to change while keeping decision
    weights fixed.
    (c) Success rates for selected systems. The VLN panels use panoramic or multi-view input, and the HM3D ObjectNav panel uses v2.}
    \label{fig:intro_overview}
\end{figure}

Generalist embodied navigation aims to support vision-and-language and
object-goal navigation across sensing conditions within a shared
system~\citep{anderson2018vision,batra2020objectnav,yokoyama2024hm3d}.
It requires understanding the surroundings, anticipating what a move may
reveal, and recognizing goal completion despite differences in camera
coverage and depth availability. Recent generalist models learn these
capabilities within unified perception-to-action policies trained across
tasks, enabling knowledge sharing and broader task
coverage~\citep{zhang2024uni,xue2026omninav,chu2026abot,zhang2026qwen}.
However, learning individual capabilities within the navigation policy
means that adding or improving them can require further training of the
navigation model (Figure~\ref{fig:intro_overview}a).

The extended mind thesis proposes that cognitive processes can extend
beyond an individual through reliable interactions with external
resources~\citep{clark1998extended}. Inspired by this view, we argue that a
navigation decision model need not master every capability itself; it can
instead use independently pretrained specialist models as external tools
to expand navigation capabilities without retraining the decision model.

We present HoloNav-0, a loosely coupled framework
that pairs a shared navigation decision model trained across tasks with
specialist foundation models as external tools
(Figure~\ref{fig:intro_overview}b). Perception, foresight, and verification
tools provide complementary information for navigation. Perception tools recover
missing depth and reconstruct panoramic context, providing compatible
observations across sensing conditions. Foresight tools generate future
observations conditioned on candidate moves, allowing the decision model
to compare their anticipated outcomes before execution. The decision
model uses the goal, observation history, and available perceptual and
predictive information to select a movement or propose stopping.
Verification tools then ground the target to validate stopping proposals
before navigation terminates.

By maintaining consistent interfaces, HoloNav-0 allows compatible tool
backends to be replaced or upgraded without updating the navigation
decision model, providing a route to benefiting from independent
advances in specialist models. Perception tools further adapt
observations across panoramic and forward-facing RGB and RGB-D
configurations, enabling the same checkpoint to operate without
sensor-specific fine-tuning.

We evaluate HoloNav-0 on vision-and-language navigation using R2R-CE and
RxR-CE~\citep{krantz2020beyond,ku2020room}, closed-vocabulary object-goal
navigation using MP3D and HM3D~\citep{chang2017matterport3d,yadav2023habitat},
and open-vocabulary object-goal navigation using
HM3D-OVON~\citep{yokoyama2024hm3d}. With one shared checkpoint and no
benchmark-specific fine-tuning, HoloNav-0 achieves competitive performance
across these task families (Figure~\ref{fig:intro_overview}c).
Tool ablations on R2R-CE and evaluations across sensor configurations
further characterize how external tools support the shared navigator.

Our contributions are threefold:
\begin{itemize}
    \item First, inspired by the extended mind thesis, we propose
    separating navigation decision-making from the acquisition of
    specialist capabilities, allowing generalist navigation capabilities
    to be jointly realized by a shared decision model and independently
    pretrained external tools.
    \item Second, we develop HoloNav-0, a loosely coupled framework that
    connects perception, foresight, and verification tools with a shared
    navigation decision model through compatible interfaces, supporting
    independent tool replacement and adaptation to different sensor
    configurations.
    \item Last, we demonstrate competitive performance across
    vision-and-language and closed- and open-vocabulary object-goal
    navigation with one checkpoint, together with adaptation across
    panoramic and forward-facing RGB and RGB-D configurations without
    sensor-specific fine-tuning.
\end{itemize}

\section{Related Work}

\subsection{Generalist Embodied Navigation}
Embodied navigation encompasses diverse tasks, including
vision-and-language navigation~\citep{anderson2018vision,lyu2026instruction} and object-goal
navigation~\citep{batra2020objectnav,yokoyama2024hm3d}.
Task-specific approaches use room-level guidance or hierarchical reasoning for image-goal
navigation~\citep{li2025regnav,li2026think} and study when to pursue or abandon goals
under an action budget~\citep{lin2026bat}. For route following, they track progress and
landmarks~\citep{wu2026dual}, maintain hierarchical memory~\citep{lyu2026himemvln},
or learn action-conditioned transitions~\citep{li2026spaact}.
This diversity has motivated a transition from task-specific systems
toward generalist navigation models.
Uni-NaVid~\citep{zhang2024uni} unifies multiple tasks through a video-based
vision-language-action policy, while NavFoM~\citep{zhang2026embodied}
extends shared navigation learning across tasks and embodiments.
OmniNav~\citep{xue2026omninav} and ABot-N0~\citep{chu2026abot} combine
high-level reasoning with trajectory generation, and
Qwen-RobotNav~\citep{zhang2026qwen} develops a scalable navigation model
through joint vision-language and trajectory training.
These efforts advance knowledge sharing and broaden the range of tasks
and environments supported by a common navigation model.
However, adding or improving a capability learned within the navigation
policy can require further training of that policy.
This raises a complementary question: once a shared navigation model
has been trained, how can it benefit from independently pretrained
specialist models without updating its decision weights?
HoloNav-0 investigates this direction by separating navigation decisions
from the acquisition of specialist capabilities, combining a shared
cross-task decision model with external perception, foresight, and
verification tools.

\subsection{Specialist Foundation Models}
Large-scale pretraining has produced models with strong capabilities in
visual geometry, controllable visual generation, and language-conditioned
localization.
Depth Anything 3~\citep{lin2025depth} recovers spatially consistent
geometry from single or multiple images, providing depth information
when direct measurements are unavailable.
Generative models offer complementary priors for views beyond the
current observation: Stable Virtual Camera
(SEVA)~\citep{zhou2025stable} synthesizes novel views under camera control,
while OmniRoam~\citep{liu2026omniroam} generates panoramic videos
conditioned on visual observations and camera trajectories.
Navigation World Models~\citep{bar2025navigation} further demonstrate
how action-conditioned predictions of future observations can support
navigation planning and the evaluation of candidate trajectories.
FS-Nav~\citep{wu2026beyond} represents future states with waypoints and route corridors
for vision-and-language navigation.
For language-conditioned localization,
LocateAnything~\citep{wang2026locateanything} grounds target descriptions
in images by predicting bounding boxes.
Together, these advances provide reusable geometric, predictive, and
grounding capabilities learned independently of a navigation policy.
HoloNav-0 draws on these capabilities through external tools that support a
shared navigation decision model.

\section{Method}
\label{sec:method}

\begin{figure}[t]
    \centering
    \includegraphics[width=\linewidth]{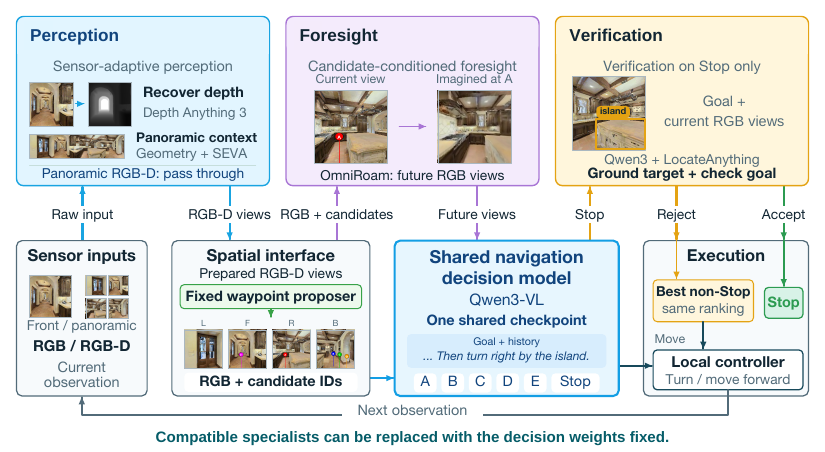}
    \caption{\textbf{Overview of HoloNav-0.} Perception tools adapt sensor inputs
    into RGB-D observations. A fixed waypoint proposer supplies feasible
    candidates, while foresight generates candidate-conditioned future views.
    The shared decision model combines RGB observations, candidate identifiers,
    the goal, history, and foresight evidence to propose a move or \textsc{Stop}.
    Verification grounds targets to validate stopping proposals. After training, compatible tools can change
    while decision weights remain fixed.}
    \label{fig:framework}
\end{figure}

\subsection{Framework and Shared Navigation Decision Model}
\label{sec:method_overview}
\label{sec:selector_action}

HoloNav-0 formulates embodied navigation as hierarchical decision making over a shared waypoint interface. At timestep $t$, the agent uses an egocentric observation $O_t$, a language goal $L$, and history $H_t$ to choose a locally reachable waypoint or \textsc{Stop}. Figure~\ref{fig:framework} summarizes the shared decision model and its external tools. HoloNav-0 expresses heterogeneous navigation goals through language. For VLN, $L$ is a route instruction. For ObjectNav and OVON, it specifies a target category. These tasks share one prompt structure and waypoint output interface and differ primarily in the instruction text. The navigation harness first converts the raw sensor stream into a heading-indexed spatial observation
\begin{equation}
    \widetilde{O}_t
    =
    \left\{
    \left(\widetilde{I}_{t}^{k},\widetilde{D}_{t}^{k},C_t^k\right)
    \right\}_{k=0}^{N-1},
    \qquad
    \alpha_k=\frac{2\pi k}{N},
\end{equation}
where $N=12$, view $0$ is aligned with the agent heading, $\alpha_k$ is the relative yaw of view $k$, and $C_t^k$ optionally records view validity.
A fixed waypoint proposer predicts a polar traversability map from $\widetilde{O}_t$ and extracts feasible candidates using circular non-maximum suppression:
\begin{equation}
    \mathcal{W}_t=\{w_t^i\}_{i=1}^{M_t},
    \qquad
    w_t^i=(\phi_t^i,d_t^i).
\end{equation}
Here, $\phi_t^i$ and $d_t^i$ denote the bearing and distance of candidate $i$. Candidate locations are projected onto four cardinal RGB views and marked with identifiers such as \texttt{A}, \texttt{B}, and \texttt{C}. These identifiers link the visual prompt and action interface to explicit spatial targets.
Rather than exposing model-specific APIs, HoloNav-0 organizes foundation capabilities through a small set of shared tool interfaces:
\begin{align}
    \mathcal{T}_{\mathrm{per}}
    &=\{\texttt{complete\_panorama},\texttt{recover\_depth}\},\\
    \mathcal{T}_{\mathrm{fore}}
    &=\{\texttt{imagine}(i)\},\\
    \mathcal{T}_{\mathrm{ver}}
    &=\{\texttt{verify}(L)\},\\
    \mathcal{T}_{\mathrm{act}}
    &=\{\texttt{move\_to}(i),\texttt{stop}()\}.
\end{align}
Perception tools normalize observations, foresight imagines candidate outcomes, verification provides target evidence, and action tools execute moves or terminate the episode. Perception is routed by the sensor configuration, and foresight can be enabled to inform waypoint selection. Compatible tool backends can be replaced through these interfaces while navigation decision weights remain unchanged.

\paragraph{Shared decision model.}
We instantiate $\mathcal{S}_{\theta}$ with Qwen3-VL~\citep{bai2025qwen3}:
\begin{equation}
    u_t
    =
    \mathcal{S}_{\theta}
    \left(L,H_t,\widetilde{O}_t^{\mathrm{RGB}},
    \mathcal{W}_t,E_t^{\mathrm{fore}}\right)
    \in
    \mathcal{W}_t\cup\{\textsc{Stop}\}.
\end{equation}
Here, $\widetilde{O}_t^{\mathrm{RGB}}$ denotes the annotated cardinal RGB views. Depth supports waypoint proposal and local control. The benchmark stop check uses RGB grounding. The history $H_t$ contains observations from previous waypoint boundaries, and $E_t^{\mathrm{fore}}$ pairs queried candidate identifiers with imagined future views when foresight is enabled and is empty otherwise. The model outputs a candidate identifier or \textsc{Stop}, rather than an open-ended plan. One checkpoint is trained with token-level supervision on mixed navigation tasks and reused across supported tasks, sensor configurations, and tool settings.

\paragraph{Action execution and stopping.}
If $u_t=w_t^i$, the harness invokes
\begin{equation}
    \texttt{move\_to}(i)
    \rightarrow
    a_{t:t+\tau}
    =\pi_{\mathrm{local}}(O_t,w_t^i).
\end{equation}
The local controller turns toward the waypoint and advances until it is reached or replanning is required, with $\tau$ denoting the execution duration. If $u_t=\textsc{Stop}$ and verification is enabled, the harness invokes \texttt{verify}(L): acceptance invokes \texttt{stop}() and ends the episode, while rejection executes the highest-scoring non-\textsc{Stop} waypoint from the same decision output. When verification is disabled, \textsc{Stop} directly ends the episode.

\subsection{Perception Tools}
\label{sec:perception_tools}

In practice, robots differ in their sensing capabilities. Perception tools enable HoloNav-0 to reuse the same navigation weights across these configurations without retraining. Panorama reconstruction provides panoramic context from limited-field-of-view observations, while depth recovery estimates missing depth. For a forward-facing RGB camera, estimated depth supplies the geometry for panorama reconstruction. Native panoramic RGB-D observations are used directly to form $\widetilde{O}_t$. Otherwise, the harness invokes only the required perception tools.

\paragraph{Panorama reconstruction.}
This tool reconstructs panoramic RGB-D observations from a single forward-facing RGB-D camera. HoloNav-0 fuses posed RGB-D observations into a colored point cloud and renders $N=12$ heading-indexed views at the current position, retaining the current front view. Geometric rendering supplies RGB, nearest-surface depth, and validity masks $C_t^k$. When coverage is insufficient, an in-place active scan captures additional views, which are fused before rendering again. SEVA~\citep{zhou2025stable} can complete residual RGB gaps using its pretrained generative prior, without changing the validity of unobserved geometry.

\paragraph{Depth recovery.}
For panoramic RGB input without measured depth, HoloNav-0 uses DA3~\citep{lin2025depth} to jointly predict relative depth maps $\{\widehat{D}_t^k\}_{k=0}^{N-1}$ from the $N$ panoramic RGB views in a single forward pass. A shared scale derived from the median valid depth across the panorama avoids independent normalization across headings. The normalized maps replace the depth inputs to the frozen waypoint predictor, while the selector retains the original RGB observations.

\subsection{Foresight Tool}
\label{sec:foresight_skill}

The foresight tool imagines the front RGB view at a candidate waypoint before execution. We instantiate $\mathcal{F}$ with OmniRoam~\citep{liu2026omniroam}, a pretrained camera-controlled panoramic video model. Let $\mathcal{V}_t$ denote the observed RGB frames used for conditioning. For candidate $w_t^i$, the harness derives a camera trajectory $\Gamma_t^i$ from the current camera pose and waypoint. OmniRoam conditions generation on $\mathcal{V}_t$ and the trajectory's motion directions and displacement scale. The interface returns
\begin{equation}
    \texttt{imagine}(i)
    \rightarrow
    \widehat{I}_{t+1}^{\,i}
    =\Pi_{\mathrm{front}}^{\,i}
    \left[\mathcal{F}(\mathcal{V}_t,\Gamma_t^i)\right],
\end{equation}
where $\Pi_{\mathrm{front}}^{\,i}$ extracts a perspective view from the generated panorama at the trajectory endpoint, aligned with the expected arrival heading. For queried candidates $\mathcal{I}_t$, the evidence
$E_t^{\mathrm{fore}}=\{(i,\widehat{I}_{t+1}^{\,i})\}_{i\in\mathcal{I}_t}$
pairs each imagined view with its waypoint index. These views provide semantic evidence for comparing candidate moves.

\subsection{Verification Tool}
\label{sec:verification_skill}

The verification tool converts the instruction into a grounding query and localizes the target in the current observation through the following interface:
\begin{equation}
    \texttt{verify}(L)
    \rightarrow
    E_t^{\mathrm{ver}}
    =\mathrm{Ground}
    \left(\widetilde{O}_t^{\mathrm{RGB}},\mathrm{TargetParser}(L)\right).
\end{equation}
A Qwen3-based parser~\citep{yang2025qwen3} extracts the target object and its descriptive constraints from $L$. LocateAnything~\citep{wang2026locateanything} then grounds the query in the current cardinal views and returns a target location and confidence.
A valid grounding confirms the proposed \textsc{Stop}.
If grounding fails, the agent select the highest-scoring waypoint to navigate to. 

\subsection{Data Collection and Training}
\label{sec:data_training}

\paragraph{Data Collection.} We collect navigation records from R2R-CE~\citep{krantz2020beyond}, RxR-CE~\citep{ku2020room}, ScaleVLN~\citep{wang2023scaling}, ObjectNav (HM3D/MP3D)~\citep{yadav2023habitat,chang2017matterport3d}, and OVON~\citep{yokoyama2024hm3d}. Native panoramic RGB-D inputs provide canonical observations for the fixed waypoint proposer during decision-model training. The simulator renders the arrival view at each queried candidate pose for Stage~1 and DAgger collection. Stage~2 uses these stored views. During closed-loop evaluation, the selected backend produces these views online using only observations available before the move and the proposed camera motion. Perception and verification tools are integrated afterward without further training of the decision model. The target is a teacher-selected candidate identifier or \textsc{Stop}, determined from the reference path or navigation goal.

\paragraph{Training Scheme.} We train the Qwen3-VL navigation decision model $\mathcal{S}_{\theta}$ in two stages, both using token-level supervision of the target action on eight NVIDIA H20 GPUs.
\textbf{Stage 1} fine-tunes the pretrained model on 1.1M standard training samples for 26 hours. Based on this model, we employ the DAgger~\citep{ross2011reduction} algorithm to collect 1.7M additional training samples.
\textbf{Stage 2} jointly trains on the combined 2.8M standard and DAgger training samples for 60 hours. This stage restarts from the original pretrained Qwen3-VL checkpoint rather than continuing from the Stage 1 weights. Supplementary Section~D.2 reports the results for each training mixture.

\section{Experiments}
\label{sec:experiments}

\begin{table}[!htb]
\centering
\setlength{\tabcolsep}{3pt}
\caption{Vision-and-language navigation on Val-Unseen. Methods are grouped by camera coverage and retain their published sensing and evaluation protocols. Our forward-facing configuration accumulates posed observations and uses active scans (4.4 per episode on average). NaVid uses the released R2R+RxR checkpoint~\citep{zhang2024navidweights}. NavMorph updates its scene memory during evaluation. Its panoramic row uses the HNR-based variant. Bold and underlining mark the best and second-best values among the listed methods in each group.}
\label{tab:vln_results}
\begin{center}
\begin{tabular*}{\linewidth}{@{\extracolsep{\fill}}lrrrrrrr@{}}
\toprule
& \multicolumn{4}{c}{R2R-CE} & \multicolumn{3}{c}{RxR-CE} \\
\cmidrule(lr){2-5}\cmidrule(l){6-8}
Method & NE$\downarrow$ & OS$\uparrow$ & SR$\uparrow$ & SPL$\uparrow$ & NE$\downarrow$ & SR$\uparrow$ & SPL$\uparrow$ \\
\midrule
\multicolumn{8}{l}{\emph{Forward-facing camera}} \\
NaVid~\citep{zhang2024navid} & 5.65 & 49.2 & 41.9 & 36.5 & 5.72 & 45.7 & 38.2 \\
Uni-NaVid~\citep{zhang2024uni} & 5.58 & 53.3 & 47.0 & 42.7 & 6.24 & 48.7 & 40.9 \\
NaVILA~\citep{cheng2024navila} & 5.22 & 62.5 & 54.0 & 49.0 & 6.77 & 49.3 & 44.0 \\
NavMorph~\citep{yao2025navmorph} & 5.75 & 56.9 & 47.9 & 33.2 & 8.85 & 30.8 & 22.8 \\
StreamVLN~\citep{wei2025streamvln} & 4.98 & 64.2 & 56.9 & 51.9 & 6.22 & 52.9 & 46.0 \\
JanusVLN~\citep{zeng2026janusvln} & 4.78 & 65.2 & 60.5 & 56.8 & 6.06 & 56.2 & 47.5 \\
AwareVLN~\citep{guo2026awarevln} & \underline{4.02} & \underline{73.5} & 65.4 & 55.1 & \textbf{3.95} & 67.6 & 56.1 \\
Qwen-RobotNav-4B~\citep{zhang2026qwen} & 4.22 & \textbf{73.6} & \textbf{66.9} & \textbf{60.5} & \underline{4.15} & \underline{71.3} & \underline{61.5} \\
Qwen-RobotNav-8B~\citep{zhang2026qwen} & 4.36 & 72.7 & 65.7 & \underline{59.6} & 4.16 & \textbf{73.4} & \textbf{63.5} \\
\midrule
\textbf{HoloNav-0} & \textbf{3.95} & \textbf{73.6} & \underline{66.8} & 58.5 & 4.67 & 66.6 & 49.2 \\
\midrule
\multicolumn{8}{l}{\emph{Panoramic or multi-view input}} \\
ETPNav~\citep{an2024etpnav} & 4.71 & 65.0 & 57.0 & 49.0 & 5.64 & 54.7 & 44.8 \\
HNR~\citep{wang2024lookahead} & 4.42 & 67.0 & 61.0 & 51.0 & 5.50 & 56.3 & 46.7 \\
NavMorph~\citep{yao2025navmorph} & 4.37 & 68.0 & 64.0 & 53.0 & 5.70 & 58.0 & 49.0 \\
NavFoM~\citep{zhang2026embodied} & 4.61 & 72.1 & 61.7 & 55.3 & 4.74 & 64.4 & 56.2 \\
NavForesee~\citep{liu2026navforesee} & 3.94 & \textbf{78.4} & 66.2 & 59.7 & 4.20 & 66.3 & 53.2 \\
ABot-N0~\citep{chu2026abot} & \underline{3.78} & 70.8 & 66.4 & \textbf{63.9} & \underline{3.83} & \underline{69.3} & \underline{60.0} \\
Qwen-RobotNav-4B~\citep{zhang2026qwen} & 3.80 & 77.2 & \underline{69.5} & \underline{63.6} & \textbf{3.80} & \textbf{75.2} & \textbf{65.0} \\
\midrule
\textbf{HoloNav-0} & \textbf{3.45} & \underline{77.5} & \textbf{71.6} & 63.1 & 4.07 & 69.2 & 54.5 \\
\bottomrule
\end{tabular*}
\end{center}
\end{table}

\subsection{Experimental Setup}
\label{sec:experiment_setup}

\paragraph{Benchmarks and metrics.}
We evaluate vision-and-language navigation on the Val-Unseen splits of
R2R-CE~\citep{krantz2020beyond} and RxR-CE~\citep{ku2020room},
closed-vocabulary ObjectNav on MP3D~\citep{chang2017matterport3d} and
HM3D~\citep{yadav2023habitat}, and open-vocabulary ObjectNav on the
Seen, Synonyms, and Unseen validation splits of
HM3D-OVON~\citep{yokoyama2024hm3d}.
For VLN, we report navigation error (NE), success rate (SR), and success
weighted by path length (SPL), with oracle success rate (OS) additionally
reported on R2R-CE. ObjectNav uses SR and SPL.
NE is measured in meters. All other metrics are percentages. The benchmark comparisons use the final Stage~2 checkpoint.
\paragraph{Implementation details.} 
We initialize the shared high-level selector from Qwen3-VL-4B-Instruct~\citep{bai2025qwen3} and use a frozen pretrained RGB-D waypoint predictor~\citep{shi2025smartway} to construct the candidate decision set.
We freeze the Qwen3-VL visual backbone while fine-tuning the complete language
model together with the visual merger/projector modules. All trainable
parameters share a peak learning rate of $2\times10^{-5}$. Training uses AdamW with
$\beta_1=0.9$, $\beta_2=0.999$, $\epsilon=10^{-8}$, weight decay $0.05$, a
linear learning-rate schedule, and a $10\%$ warmup ratio. We train on eight H20 GPUs with four examples per GPU and four gradient-accumulation steps, giving an effective batch size of 128. We use bfloat16, FlashAttention-2,
DeepSpeed ZeRO-2, and gradient checkpointing.

\subsection{Benchmark Results}
\label{sec:main_results}

\paragraph{Vision-and-language navigation.}
With panoramic input, HoloNav-0 achieves 71.6\% SR on R2R-CE and 69.2\% on RxR-CE (Table~\ref{tab:vln_results}). Compared with the panoramic Qwen-RobotNav-4B, HoloNav-0 gains 2.1 percentage points in R2R-CE SR. On RxR-CE, Qwen-RobotNav-4B reports higher SR and SPL (75.2 and 65.0). With the same decision weights and a forward-facing camera, HoloNav-0 attains 66.8\% and 66.6\% SR on the two benchmarks. Camera coverage alone does not control for differences in depth, pose access, active sensing, or training data across systems.
\begin{table}[!htb]
\centering
\setlength{\tabcolsep}{4pt}
\caption{Closed-vocabulary object-goal navigation. Both Qwen-RobotNav variants, LightNav-0 and HoloNav-0 report HM3D v2 results. The other HM3D scores are from v1, except for CompassNav, whose version is unspecified. Bold and underlining mark the highest and second-highest reported values in each column, respectively.}
\label{tab:objectnav_results}
\begin{center}
\begin{tabular*}{\linewidth}{@{\extracolsep{\fill}}lrrrr@{}}
\toprule
& \multicolumn{2}{c}{MP3D} & \multicolumn{2}{c}{HM3D} \\
\cmidrule(lr){2-3}\cmidrule(l){4-5}
Method & SR$\uparrow$ & SPL$\uparrow$ & SR$\uparrow$ & SPL$\uparrow$ \\
\midrule
VLFM~\citep{yokoyama2024vlfm} & 36.4 & 17.5 & 52.5 & 30.4 \\
OpenFMNav~\citep{kuang2024openfmnav} & 37.2 & 15.7 & 52.5 & 24.1 \\
SG-Nav~\citep{yin2024sg} & 40.2 & 16.0 & 54.0 & 24.9 \\
TriHelper~\citep{zhang2024trihelper} & -- & -- & 56.5 & 25.3 \\
UniGoal~\citep{yin2025unigoal} & 41.0 & 16.4 & 54.5 & 25.1 \\
WMNav~\citep{nie2025wmnav} & 45.4 & 17.2 & 58.1 & 31.2 \\
CogNav~\citep{cao2025cognav} & 46.6 & 16.1 & 72.5 & 26.2 \\
Uni-NaVid~\citep{zhang2024uni} & -- & -- & 73.7 & 37.1 \\
CompassNav~\citep{li2026compassnav} & 42.0 & 17.5 & 56.6 & 27.6 \\
Qwen-RobotNav-4B~\citep{zhang2026qwen} & 52.2 & 16.0 & \underline{75.6} & 30.6 \\
Qwen-RobotNav-8B~\citep{zhang2026qwen} & 48.8 & 17.7 & 71.2 & 33.0 \\
LightNav-0~\citep{wang2026lightnav} & \underline{53.3} & \underline{21.2} & \textbf{77.2} & \underline{41.5} \\
\midrule
\textbf{HoloNav-0} & \textbf{55.3} & \textbf{28.3} & 75.2 & \textbf{45.6} \\
\bottomrule
\end{tabular*}
\end{center}
\end{table}

\paragraph{Closed-vocabulary object-goal navigation.}
HoloNav-0 reaches 55.3\% SR and 28.3\% SPL on MP3D (Table~\ref{tab:objectnav_results}). On HM3D v2, LightNav-0 has the highest listed SR (77.2\%). HoloNav-0 has a lower SR (75.2\%) and the highest listed SPL (45.6\%).
\begin{table}[!htb]
\centering
\setlength{\tabcolsep}{3pt}
\caption{Open-vocabulary ObjectNav results on the HM3D-OVON validation
splits~\citep{yokoyama2024hm3d}. Bold and underlining denote the best
and second-best results, respectively, in each column;
-- denotes an unreported result.
$\dagger$ marks baseline results reported with depth or odometry.
OVSegDT and JanusVLN use monocular RGB input. JanusVLN reports only
the Unseen split.}
\label{tab:ovon_results}
\begin{center}
\begin{tabular*}{\linewidth}{@{\extracolsep{\fill}}lcccccc@{}}
\toprule
& \multicolumn{2}{c}{Val Seen} & \multicolumn{2}{c}{Val Synonyms} & \multicolumn{2}{c}{Val Unseen} \\
\cmidrule(lr){2-3}\cmidrule(lr){4-5}\cmidrule(lr){6-7}
Method & SR$\uparrow$ & SPL$\uparrow$ & SR$\uparrow$ & SPL$\uparrow$ & SR$\uparrow$ & SPL$\uparrow$ \\
\midrule
VLFM$^\dagger$~\citep{yokoyama2024vlfm} & 35.2 & 18.6 & 32.4 & 17.3 & 35.2 & 19.6 \\
DAgRL+OD$^\dagger$~\citep{yokoyama2024hm3d} & 38.5 & 21.1 & 39.0 & 21.4 & 37.1 & 19.8 \\
MTU3D$^\dagger$~\citep{zhu2025move} & 55.0 & 23.6 & 45.0 & 14.7 & 40.8 & 12.1 \\
Uni-NaVid~\citep{zhang2024uni} & 41.3 & 21.1 & 43.9 & 21.8 & 39.5 & 19.8 \\
OVSegDT~\citep{zemskova2025ovsegdt} & 43.6 & 20.1 & 40.1 & 17.9 & 44.7 & 20.6 \\
JanusVLN~\citep{zeng2026janusvln} & -- & -- & -- & -- & 44.9 & 31.7 \\
NavFoM (4-view)~\citep{zhang2026embodied} & 40.1 & 27.1 & 45.4 & 32.6 & 45.2 & \underline{31.9} \\
LightNav-0~\citep{wang2026lightnav} &55.3 &31.2 &54.6 &29.6 &47.0 &24.2 \\
ABot-N0~\citep{chu2026abot} & 55.3 & \underline{32.1} & 55.4 & \underline{33.2} & \underline{54.0} & 30.5 \\
Qwen-RobotNav-4B~\citep{zhang2026qwen} & \underline{57.7} & 24.4 & \textbf{60.1} & 25.1 & 53.1 & 20.9 \\
Qwen-RobotNav-8B~\citep{zhang2026qwen} & 56.1 & 28.5 & 57.8 & 28.8 & 51.2 & 24.0 \\
\midrule
\textbf{HoloNav-0} & \textbf{60.4} & \textbf{36.4} & \underline{59.2} & \textbf{35.8} & \textbf{56.7} & \textbf{33.2} \\
\bottomrule
\end{tabular*}
\end{center}\par
\end{table}

\paragraph{Open-vocabulary object-goal navigation.}
HoloNav-0 achieves the highest listed SPL on all three HM3D-OVON splits and the
highest listed SR on Seen and Unseen (Table~\ref{tab:ovon_results}).
On Unseen, it exceeds ABot-N0 by 2.7 percentage points in both SR and
SPL. On Synonyms, its SR is lower than Qwen-RobotNav-4B
(59.2\% vs.\ 60.1\%). Together with the VLN and closed-vocabulary
results, these comparisons support competitive performance across tasks
with one shared decision checkpoint. The following ablations examine
individual tools within this complete system.

\subsection{Ablation Study}
\label{sec:ablations}

\begin{table}[!htb]
\centering
\setlength{\tabcolsep}{3pt}
\caption{Tool ablations and sensor adaptation on R2R-CE Val-Unseen with the Stage~1 checkpoint. Panel~(a) uses native panoramic RGB-D. Panel~(b) disables foresight and verification in all four sensor settings. Front denotes a single forward-facing camera. A dash denotes a disabled or unnecessary tool.}
\label{tab:skill_ablation}
\begin{center}
\begin{tabular*}{\linewidth}{@{\extracolsep{\fill}}lcccccccc@{}}
\toprule
& \multicolumn{4}{c}{Tools} & \multicolumn{4}{c}{R2R-CE Val-Unseen} \\
\cmidrule(lr){2-5}\cmidrule(lr){6-9}
Setting & \shortstack{Depth\\recovery} & \shortstack{Panorama\\reconstruction} & Foresight & Verification & NE$\downarrow$ & OS$\uparrow$ & SR$\uparrow$ & SPL$\uparrow$ \\
\midrule
\multicolumn{9}{l}{\textit{(a) Foresight and verification: panoramic RGB-D}} \\
Both tools & -- & -- & $\checkmark$ & $\checkmark$ & 3.87 & 72.60 & 69.03 & 63.53 \\
w/o verification & -- & -- & $\checkmark$ & -- & 4.00 & 72.32 & 68.62 & 64.17 \\
w/o foresight & -- & -- & -- & $\checkmark$ & 4.05 & 71.17 & 66.47 & 62.12 \\
w/o both & -- & -- & -- & -- & 4.23 & 70.85 & 66.01 & 61.88 \\
\midrule
\multicolumn{9}{l}{\textit{(b) Sensor adaptation: foresight and verification disabled}} \\
Panoramic RGB-D & -- & -- & -- & -- & 4.23 & 70.85 & 66.01 & 61.88 \\
Panoramic RGB & $\checkmark$ & -- & -- & -- &  4.12 &71.29  &65.91  &61.60  \\
Front RGB-D & -- & $\checkmark$ & -- & -- & 4.46 & 68.41 & 62.70 & 58.66 \\
Front RGB & $\checkmark$ & $\checkmark$ & -- & -- & 4.95 & 62.28 & 57.37 & 52.32 \\
\bottomrule
\end{tabular*}
\end{center}\par
\end{table}

Table~\ref{tab:skill_ablation} reports tool ablations and four sensor configurations on R2R-CE Val-Unseen, with the same waypoint proposer, local controller, evaluation episodes, and action budgets.

\paragraph{Foresight and verification.}
Table~\ref{tab:skill_ablation} shows that foresight alone improves R2R-CE SR by 2.61 percentage points, compared with 0.46 for verification. Combining both achieves the highest SR of 69.03\%, but lowers SPL relative to foresight alone, indicating a trade-off between task completion and path efficiency.
With Stage~1 on HM3D-OVON Unseen, verification alone raises SR from 49.67\% to 51.23\%, while foresight alone gives 49.72\%. The final Stage~2 checkpoint shows the same task dependence. On R2R-CE, SR is 69.1\% with both tools off, 70.8\% with foresight alone, 70.5\% with verification alone, and 71.6\% with both. On HM3D-OVON Unseen, the corresponding rates are 55.0\%, 55.4\%, 56.6\%, and 56.7\%. These are point estimates. In particular, the 0.1-point SR difference between verification alone and both tools on HM3D-OVON does not establish an additional benefit from foresight.

\paragraph{Specialist replacement.}
With the Stage~2 policy fixed and foresight disabled, replacing LocateAnything with GPT-6 Astra raises SR by 1.7 percentage points on R2R-CE and 1.8 points on HM3D-OVON Val-Unseen. Supplementary Table~S1 reports the full comparison of foresight backends and stop verifiers.

\paragraph{Adaptation to sensor settings.}
With foresight and verification disabled (Table~\ref{tab:skill_ablation}(b)), replacing sensed depth with DA3~\citep{lin2025depth} predictions under panoramic RGB input reduces SR and SPL by only 0.10 and 0.28 percentage points, respectively, largely preserving navigation performance. With front-facing RGB-D input, geometric panorama reconstruction and SEVA~\citep{zhou2025stable} completion enable the same checkpoint to operate, but SR remains 3.31 points below the panoramic RGB-D reference. Removing depth as well widens this gap to 8.64 points. Thus, perception tools support all four sensor configurations without sensor-specific fine-tuning, but do not fully compensate for limited visual coverage.

\subsection{Qualitative Results}
\label{sec:qualitative_results}

\begin{figure}[!htbp]
    \centering
    \includegraphics[width=\linewidth]{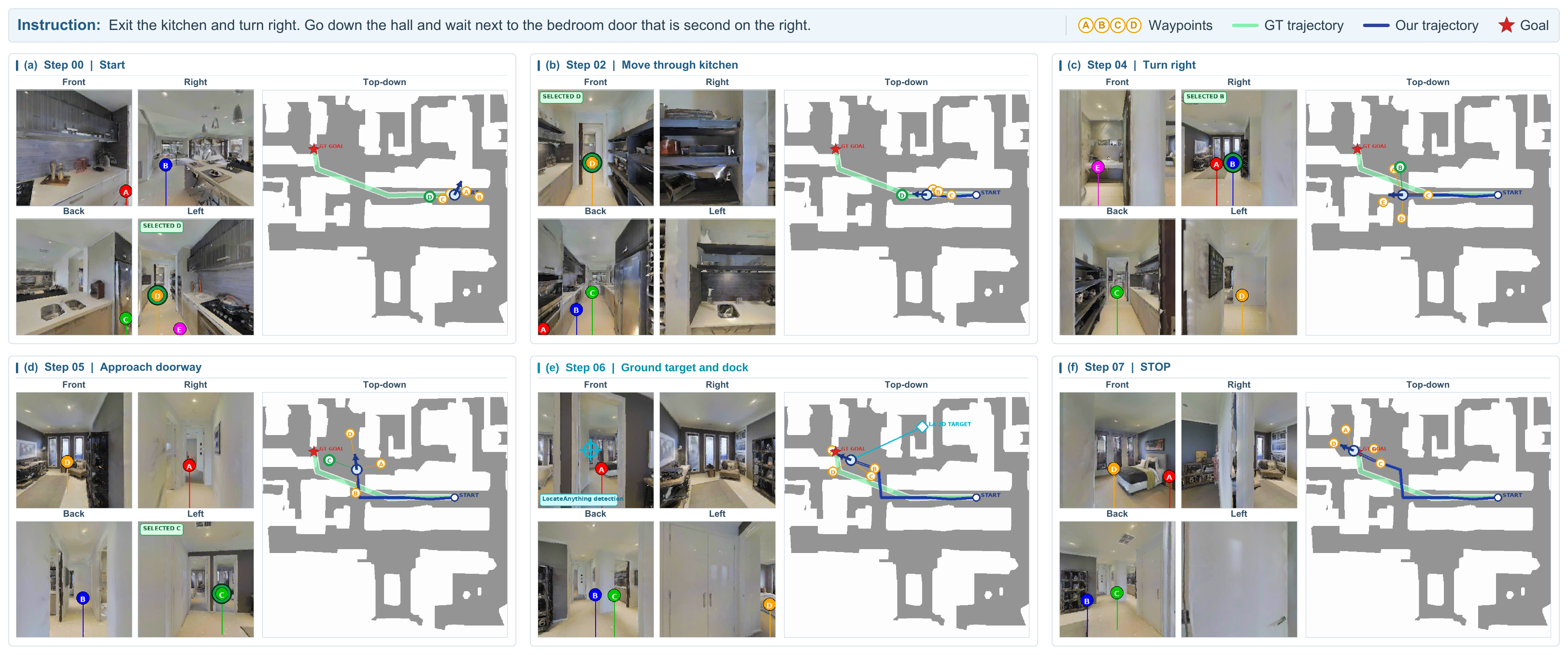}
    \caption{\textbf{Instruction following with optional docking enabled.} Six selected steps show the directional RGB observations and a top-down visualization. Lettered markers denote candidate waypoints. Green and blue lines indicate the reference and executed trajectories, and the red star marks the goal. In panel~(e), the cyan cross marks the image detection and the diamond marks the estimated 3D target for docking.}
    \label{fig:qualitative_navigation}
\end{figure}

Figure~\ref{fig:qualitative_navigation} shows an R2R-CE rollout with optional docking enabled. The agent leaves the kitchen, turns right, and follows the hallway to the second bedroom door on the right. At step~6, the verifier grounds the target and the agent docks, reducing the goal distance from 1.037\,m to 0.215\,m before stopping at step~7. Docking is disabled in all benchmark evaluations. When stop verification is enabled, it uses RGB grounding.

\section{Conclusion}
\label{sec:conclusion}

HoloNav-0 trains one navigation policy to select waypoints and propose stopping for both instruction following and object search. Specialists can change the visual evidence available to those decisions without requiring another round of policy training. At the final checkpoint, foresight alone yields the larger success-rate gain on R2R-CE, while stop verification alone yields the larger gain on HM3D-OVON. The sensor study identifies a limit to this flexibility. With foresight and verification disabled on R2R-CE, estimated depth largely preserves success under panoramic input, but reconstructing the panorama from forward-facing observations leaves a larger gap. The experiments cover simulated indoor environments with predefined rules for invoking tools. Testing those rules on physical robots and learning when an additional view is worth acquiring remain open questions.

\subsection*{AI Use Statement}
We used generative AI tools to assist with drafting and revising the manuscript, discussing how to present the work, identifying and organizing literature and comparison baselines, and preparing scientific figures. The authors designed the method, implemented the research code, conducted the experiments, and take responsibility for the text, references, and figures.

\clearpage
\bibliographystyle{iclr2027_conference}
\bibliography{references}
\end{document}

%% file: math_commands.tex
\usepackage{amsmath,amsfonts,bm}

\def\eqref#1{equation~\ref{#1}}

\def\1{\bm{1}}

\DeclareMathAlphabet{\mathsfit}{\encodingdefault}{\sfdefault}{m}{sl}
\SetMathAlphabet{\mathsfit}{bold}{\encodingdefault}{\sfdefault}{bx}{n}

